\documentclass[11pt,a4paper]{article}
\usepackage[hyperref]{emnlp-ijcnlp-2019}
\usepackage{times}
\usepackage{latexsym}
\usepackage{helvet}  
\usepackage{courier}  
\usepackage{url}  
\usepackage{graphicx}  
\usepackage{comment}

\usepackage{url}

\usepackage{url}
\usepackage{graphicx}
\usepackage{paralist}
\usepackage{amsmath}
\usepackage{amssymb}
\usepackage{subfigure}
\usepackage{booktabs}
\usepackage{multirow}
\usepackage{float}
\usepackage[ruled,noend,linesnumbered]{algorithm2e}
\usepackage{enumitem}
\restylefloat{table}
\usepackage{color}
\usepackage{tabularx}
\usepackage{array}
\usepackage{lscape}
\usepackage{amsmath}
\usepackage{slashbox}

\usepackage{mwe}
\usepackage{kantlipsum}
\usepackage{listings}

\usepackage{amsmath,amsfonts,bm}

\def\eqref#1{equation~\ref{#1}}

\def\1{\bm{1}}

\DeclareMathAlphabet{\mathsfit}{\encodingdefault}{\sfdefault}{m}{sl}
\SetMathAlphabet{\mathsfit}{bold}{\encodingdefault}{\sfdefault}{bx}{n}

\usepackage{hyperref}
\usepackage{url}

\usepackage{todonotes}
\makeatletter
\newcommand*\iftodonotes{\if@todonotes@disabled\expandafter\@secondoftwo\else\expandafter\@firstoftwo\fi}  
\makeatother

\aclfinalcopy 

\title{Robust Text Classifier on Test-Time Budgets}

\author{
\bf Md Rizwan Parvez \\
University of California Los Angeles \\
 {\tt rizwan@cs.ucla.edu} \\ 
 \\ \bf Kai-Wei Chang \\
University of California Los Angeles \\
 {\tt kwchang@cs.ucla.edu} \\
  \\
   \And Tolga Bolukbasi
\\ Boston University  \\
 {\tt tolgab@bu.edu} \\ 
 \\
{\bf Venkatesh Saligrama}\\
Boston University \\
 {\tt srv@bu.edu} \\
}
\date{}
\begin{document}
\maketitle
{
\begin{abstract}
We design a generic framework for learning a robust text classification model that achieves high accuracy under different selection budgets  (a.k.a selection rates) at test-time. We take a different approach from existing methods and learn to dynamically filter a large fraction of unimportant words by a low-complexity selector such that any high-complexity classifier only needs to process a small fraction of text, relevant for the target task. To this end, we propose a data aggregation method for training the classifier, allowing it to achieve competitive performance on fractured sentences. On four
benchmark text classification tasks, we demonstrate that the framework gains consistent speedup with little degradation in accuracy on various selection budgets.
\end{abstract}
\section{Introduction}
Recent advances in deep neural networks (DNNs) have achieved high accuracy on many text classification tasks.
These approaches process the entire text and encode words and phrases in order to perform target tasks. While these models realize high accuracy, the computational time scales linearly with the size of the documents, which can be slow for a long document.
In this context, various approaches based on modifying the  RNN or LSTM architecture have been proposed to speed up the process \cite{skim-rnn,learning-to-skim}. However, the processing in these models is still fundamentally sequential and needs to operate on the whole document which limits the computational gain.   
\begin{figure}
\centering
\includegraphics[width=7.7cm, height=4.53570425cm]{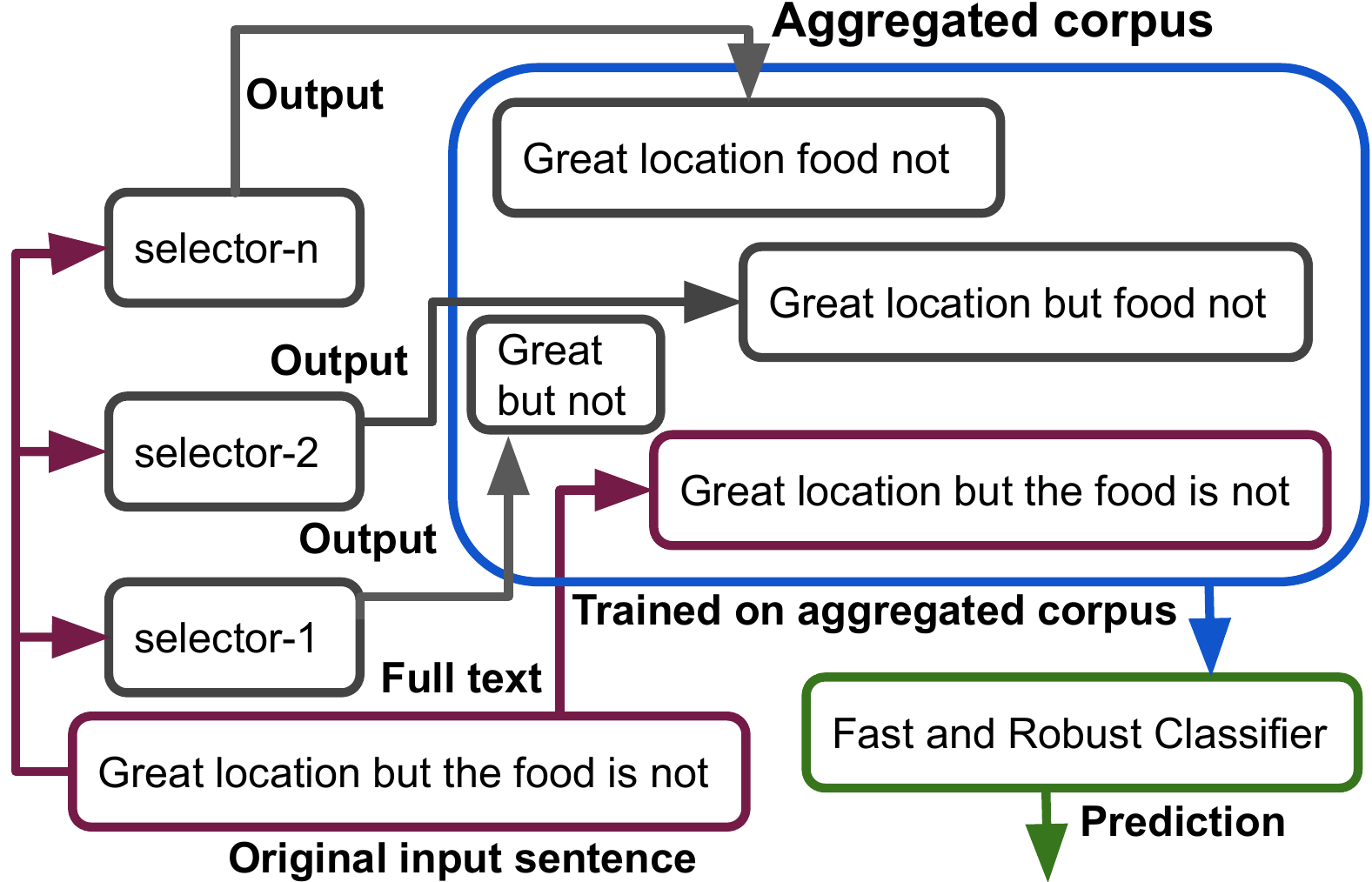}
\caption{\footnotesize { Our proposed framework. Given a selection rate, a \textit{selector} is designed to select relevant words and pass them to the \textit{classifier}. To make the classifier robust against fractured sentences, we aggregate outputs from different {\it selectors} and train the {\it classifier} on the aggregated corpus.}
\vspace{-0.2cm}
}
  \label{fig:model}
\end{figure}
In contrast to previous approaches, we propose a novel framework for efficient text classification on long documents that mitigates sequential processing. The framework consists of a \textit{selector} and a \textit{classifier}. Given a selection budget as input, the \textit{selector} performs a coarse one-shot selection deleting unimportant words and pass the remainder 
to the \textit{classifier}. 
The \textit{classifier} then takes the sentence fragments as an input and performs the target task. 
Figure ~\ref{fig:model} illustrates the procedure. 
This framework is general and agnostic to the architecture of the downstream {\it classifier} (e.g., RNN, CNN, Transformer). 

However, three challenges arise.
First, to build a computationally inexpensive system, the \textit{selector} must have negligible overhead. We adopt two effective yet simple architectures to design {\it selector}s based on word embeddings and bag-of-words. Second, training multiple distinct models for different budgets is unfeasible in practice, especially when model size is large. Hence, our goal is to learn a single {\it classifier} that can adapt to the output of any {\it selector} operating at any budget. Consequently, this {\it classifier} must be robust so that it can achieve consistent performance with different budgets. Third, the input to the \textit{classifier} in our framework is a sequence of fractured sentences which is incompatible with a standard {\it classifier} that trained on the  full texts, causing its performance degrades significantly. 
One potential but unfeasible solution is to train the {\it classifier} with a diverse collection of sentence fragments which is combinatorially numerous. Another approach is to randomly blank out text (a.k.a. blanking-noise), leads to marginalized feature distortion~\citep{maaten2013learning} but this also leads to poor accuracy as DNNs leverage word combinations, sentence structure, which this approach does not account for.
To mitigate this problem, we propose a data aggregation framework that augments the training corpus with outputs from {\it selector}s at different budget levels. By training the {\it classifier} on the aggregated {\it structured} blank-out text, the {\it classifier} learns to fuse fragmented sentences into a feature representation that mirrors the representation obtained on full sentences and thus realizes high-accuracy. 
We evaluate our approach through comprehensive experiments on real-world datasets.~\footnote{Our source code is available at: \\ \url{https://github.com/uclanlp/Fast-and-Robust-Text-Classification}}

\section{Related Work}
Several approaches have been proposed to speed up the DNN in test time~\cite{Kilian,choi}.  
LSTM-jump~\citep{learning-to-skim} learns
to completely skip words deemed to be irrelevant and skim-RNN~\citep{skim-rnn} uses a low-complexity LSTM to skim words rather than skipping. Another version of LSTM-jump, LSTM-shuttle~\cite{speed-emnlp18} first skips a number of words, then goes backward to recover lost information by reading some words skipped before. All these approaches require to modify the architecture of the underlying classifier and cannot easily extend to another architecture. 
In contrast, we adopt existing
{\it classifier}
architectures (e.g., LSTM, BCN~\citep{bcn}) 
and propose a meta-learning algorithm to train the model.
Our framework is generic and a {\it classifier} can be viewed as a black-box.
Similar to us, \citet{tao-lei} propose a \textit{selector}-\textit{classifier} framework to find text snippets as justification for text classification but their {\it selector} and {\it classifier} have similar complexity and require similar processing times; therefore, it is not suitable for computation gain.  
Various feature selection approaches~\citep{chandrashekar2014survey} have been discussed in literature. For example, removing predefined stop-words (see Appendix
A), attention based models ~\cite{bahdanau, luong}, feature subspace selection methods (e.g., PCA), and applying the L1 regularization  (e.g., Lasso~\citep{lasso1} or Group Lasso~\citep{faruqui2015sparse}, BLasso~\citep{gao}). However, these approaches either cannot obtain sparse features or cannot straightforwardly be applied to speed up a DNN {\it classifier}. 
Different from ours, \citet{ viola2001robust,trapeznikov2013supervised,karayev2013dynamic,xu2013cost,kusner2014feature,bengio2015conditional,leroux2017cascading, zhu19,NIPS2017_7058,pmlr-v70-bolukbasi17a} focus on gating various components of existing networks.
Finally, aggregating data or models has been studied under different contexts 
(e.g., in context of reinforcement learning \citep{daggar}, Bagging models~\citep{breiman1996bagging},  etc.) while we aggregate the data output from {\it selectors} instead of models.

\section{Classification on a Test-Time Budget}
\label{sec:framework}
Our goal is to build a robust {\it classifier} along with a suite of {\it selectors} to achieve good performance with consistent speedup under different selection budgets at test-time. Formally, a {\it classifier} $C(\hat{x})$ takes a word sequence $\hat{x}$ and predicts the corresponding output label $y$, and a {\it selector} $S_b(x)$ with selection budget $b$ takes an input word sequence $x=\{w_1, w_2, \ldots, w_N\}$ and generates a binary sequence $S_b(x) = \{z_{w_{1}}, z_{w_{2}}, \ldots, z_{w_{N}}\}$ where $z_{w_{k}} \in \{0,1\}$ represents if the corresponding word 
$w_k$
is selected or not. 
We denote the sub-sequence of words generated after filtering by the {\it selector} as $I\big(x, S_b(x)\big) = \left\{{w_{k}: z_{w_{k}}=1, \forall w_k \in x}\right\}$. We aim to train a {\it classifier} $C$ and the {\it selector} $S_b$ such that $I\big(x, S_b(x)\big)$ is sufficient to make accurate prediction on the output label (i.e., $C\Big(I\big(x, S_b(x)\big)\Big)\approx C(x)$).
 The selection budget (a.k.a selection rate) $b$ is controlled by the hyper-parameters of the \textit{selector}.
 Higher budget often leads to higher accuracy and longer test time.
\subsection{Learning a {\it \textbf{Selector}}}
\label{sec:learn_selctor}
We propose two simple but efficient {\it selector}s.
\noindent{\bf Word Embedding (WE) \textit{selector}.}
\label{diff_models}
We consider a parsimonious word-selector using word embeddings (e.g., GloVe~\cite{glove}) as features to predict important words. We assume the informative words can be identified independently and model the probability that a word $w_k$ is selected by $P(z_{w_{k}}=1|w_k)  =\sigma(\theta_S^T\vec{w}_{k})$,
where $\theta_S$ is the model parameters of the {\it selector} $S_b$, $\vec{w}_{k}$ is the corresponding word vector, and $\sigma$ is the sigmoid function. 
As we do not have explicit annotations about which words are important, we train the {\it selector} $S_b$ along with a {\it classifier} $C$ in an end-to-end manner following  \citet{tao-lei}, and an L1-regularizer is added to control the sparsity (i.e., selection budget) of $S_b(x)$.

\noindent{\bf{Bag-of-Words \textit{selector}.}}
\label{sec:bag-of-words}
We also consider using an L1-regularized linear model~\citep{elasticnet,ng_L1,kai_L1} with bag-of-words features to identify important words. 
In the bag-of-words model, for each document $x$, we construct a feature vector $\vec{x}\in \{0,1\}^{|V|}$, where $|V|$ is the size of the vocabulary. Each element of the feature vector $\vec{x}_w$  represents if a specific word $w$ appearing in the document $x$. Given a training set $\mathcal{X}$, the linear model optimizes the L1-regularized task loss. For example, in case of a binary classification
task (output label $y\in \{1,-1\}$),
\begin{equation*}
\label{eq:log-reg}
    \begin{split}
    J(x_t, y_t) &= \log \left({1+\exp(- y_t\theta^T \vec{x_t}})\right)    \\
    \theta^* &= 
\arg \min\nolimits_{\theta} \sum_{(x_t,y_t) \in \mathcal{X}} J(x_t, y_t) + \frac{1}{b} \| \theta \|_1 ,
    \end{split}
\end{equation*}
where $\theta \in R^{|V|}$ is a weight vector to be learned, $\theta_{w}$ corresponds to word $w \in V$,
and $b$ is a hyper-parameter controlling the sparsity of $\theta^*$ (i.e., selection budget). 
The lower the budget $b$ is, the sparser the selection is.
Based on the optimal $\theta^*$, we construct a {\it selector} that picks word $w$ if the corresponding $\theta^*_w$ is non-zero. Formally, the bag-of-words {\it selector} outputs 
$S_b(x)  = \{ \delta(\theta_w \neq 0) : w \in x\},$
where $\delta$ is an
indicator function.

\renewcommand{\tabcolsep}{2.2pt}
{
\begin{table*}[t]
\begin{minipage}{\textwidth}
\centering
\resizebox{\textwidth}{!}{%
\small
\begin{tabular}{l|l@{ }c@{ }c@{ }r|l@{ }c@{ }c@{ }r|l@{ }c@{ }c@{ }r|l@{ }c@{ }c@{ }r} 
 \toprule
 {\bf Model}  & \multicolumn{4}{c}{\bf SST-2} &
  \multicolumn{4}{c}{\bf IMDB} & \multicolumn{4}{c}{\bf AGNews}  & \multicolumn{4}{c}{\bf Yelp}   \\ 
  &  acc. & selection(\%) & time & speedup &  acc. & selection(\%) & time & speedup  &  acc.  & selection(\%)  & time & speedup &  acc.  & selection(\%)   & time & speedup\\
  
    
    \toprule
{\it Baseline}  &  {\bf 85.7} & 100 & 9 & 1x   & 91.0 & 100 & 1546& 1x & 92.3 & 100 & 59 & 1x &  {\bf 66.5} & 100 & 3487 & 1x \\
Bag-of-Words  &  78.8 & 75 & 5.34 & {1.7x}  & 91.5 & 91 & 1258 & {\bf 1.2x} & 92.9 & 97 & 48 & 1.2x &  59.7 & 55 &  2325 & {\bf 1.6x} \\
\midrule
     
    Our framework  & {82.6} & 65 & 4.6 & \textbf{2x}  & {92.0} & 91 & 1297  & {\bf 1.2x} & {93.1} & 91 & 46  & \textbf{1.3x}  &  {64.8} & 55 & 2179  & \textbf{1.6x} \\ 
 &  {85.3}  & 0 & 9 & 1x  & \textbf{92.1} & 0 & 1618 & 1x & {\bf 93.2} & 0 & 57 & 1x & 66.3 & 0 & 3448 & 1x \\
 \bottomrule
 
 \end{tabular}
 }
 \caption{\footnotesize{Accuracy and speedup on the test datasets. Test-times are measured in seconds. The speedup rate is calculated as the running time of a model divided by the running time of the corresponding baseline.
For  our framework, top row denotes the best speedup and the bottom row denotes the best test accuracy achieved. 
Overall best accuracies and best speedups are boldfaced. 
Our framework achieves accuracies better than baseline with a speedup of {\bf 1.2x} and {\bf 1.3x} on IMDB, and AGNews respectively. With same or higher speedup, our accuracies are much better than Bag-of-Words. 
}
}
\label{tab-bcnresult}
\end{minipage}

\end{table*}
}

\subsection{The Data Aggregation Framework}
\label{sec:dagger}
In order to learn to fuse fragmented sentences
into a robust feature representation, we propose to train the {\it classifier} on the aggregated corpus of structured blank-out texts.

Given a set of training data  $\mathcal{X} = \{(x_1,y_1),..,(x_t,y_t),..,(x_m,y_m)\}$, we assume we have a set of selectors $\mathcal{S} = \{S_b\}$ with different budget levels trained by the framework discussed in Section \ref{sec:learn_selctor}.
To generate an aggregated corpus, we first apply each {\it selector} $S_b \in \mathcal{S}$ on the training set, and generate corresponding blank-out corpus 
 $\mathcal{I}(\mathcal{X},S_b) =  \left\{{I \big ( x_t, S_b(x_t) \big), \forall  x_t \in \mathcal{X}}\right\}$.
 Then, we create a new corpus by aggregating the blank-out corpora: $\mathcal{T} = \bigcup_{S_b\in \mathcal{S}} \mathcal{I}(\mathcal{X},S_b)$.\footnote{Note that, the union operation is used just to aggregate the train instances which does not hinder the model training 
(e.g., discrete variables).} Finally, we train the {\it classifier} $C_\mathcal{T}$ on the aggregated corpus $\mathcal{T}$. As $C_\mathcal{T}$ is trained on documents with distortions, it learns to make predictions with different budget levels. 
 The training procedure is summarized in Algorithm 1. In the following, we discuss two extensions of our data aggregation framework.

First, the blank-out data can be generated from different classes of {\it selectors} with different features or architectures. Second, the blank-out and selection can be done in phrase or sentence level. Specifically, if phrase boundaries are provided, a phrase-level aggregation can avoid a \textit{selector} from breaking compound nouns or meaningful phrases (e.g., ``Los Angeles'', ``not bad``). Similarly, for multi-sentenced documents, we can enforce the \textit{selector} to pick a whole sentence if any word in the sentence is selected.

\begin{algorithm}[t]
\DontPrintSemicolon
\KwIn{Training corpus $\mathcal{X}$, a set of \textit{selectors} with different budget levels  $\mathcal{S} = \{S_b\}$, {\it classifier} class $C$
    }
\KwOut{A robust \textit{classifier} $C_\mathcal{T}$}
\vspace{3pt}
 Initialize the aggregated corpus: $\mathcal{T} \gets \mathcal{X}$ \;
 \For{$S_b \in \mathcal{S}$} {    
    $S_b \gets $ Train a selector $S_b \in S$ with budget level $b$ on $\mathcal{X}$ 
    \;
    Generate a blank-out dataset $\mathcal{I}(\mathcal{X},S_b)$\;
    Aggregate data: $\mathcal{T} \gets \mathcal{T} \cup \mathcal{I}(\mathcal{X},S_b)$\;
  }
 \vspace{5pt}
  $C_\mathcal{T} \gets$ Train a {\it classifier} $C$ on
  $\mathcal{T}$\;  
  \Return $C_\mathcal{T}$ \;
 \caption{\textbf{\small Data Aggregated Training Schema}}
 \label{algo-1}
\end{algorithm}

\begin{table}
\vspace{0.22cm}
\resizebox{1.0\columnwidth}{!}{
\centering
\renewcommand{\tabcolsep}{2pt}
 \begin{tabular}{l|l|l|c|c} 
 \toprule
 Dataset  & \#class & Vocabulary & Size (Train/Valid/Test)  & Avg. Len   \\ 
 \toprule
  SST & 2  &  13,750 & 6,920/872/1,821  & 19 \\
IMDB & 2 & 61,046 & 21,143/3,857/25,000 & 240 \\
  AGNews & 4 &  60,088 & 101,851/18,149/7,600 & 43 \\
 Yelp & 5 &  1,001,485 & 600k/50k/50k & 149 \\
 \bottomrule
 \end{tabular}
 \caption{\footnotesize Dataset statistics.}
  \label{data}
 }
\end{table}

\begin{figure*}[h]
\setlength{\abovecaptionskip}{-4pt}
\vspace{-0.4cm}
  \centering
    \begin{tabular}{c@{}c@{}}
    \hspace{-0.651cm}
  \subfigure[IMDB]{\
        \includegraphics[width=0.3740\linewidth, height=0.277\linewidth]{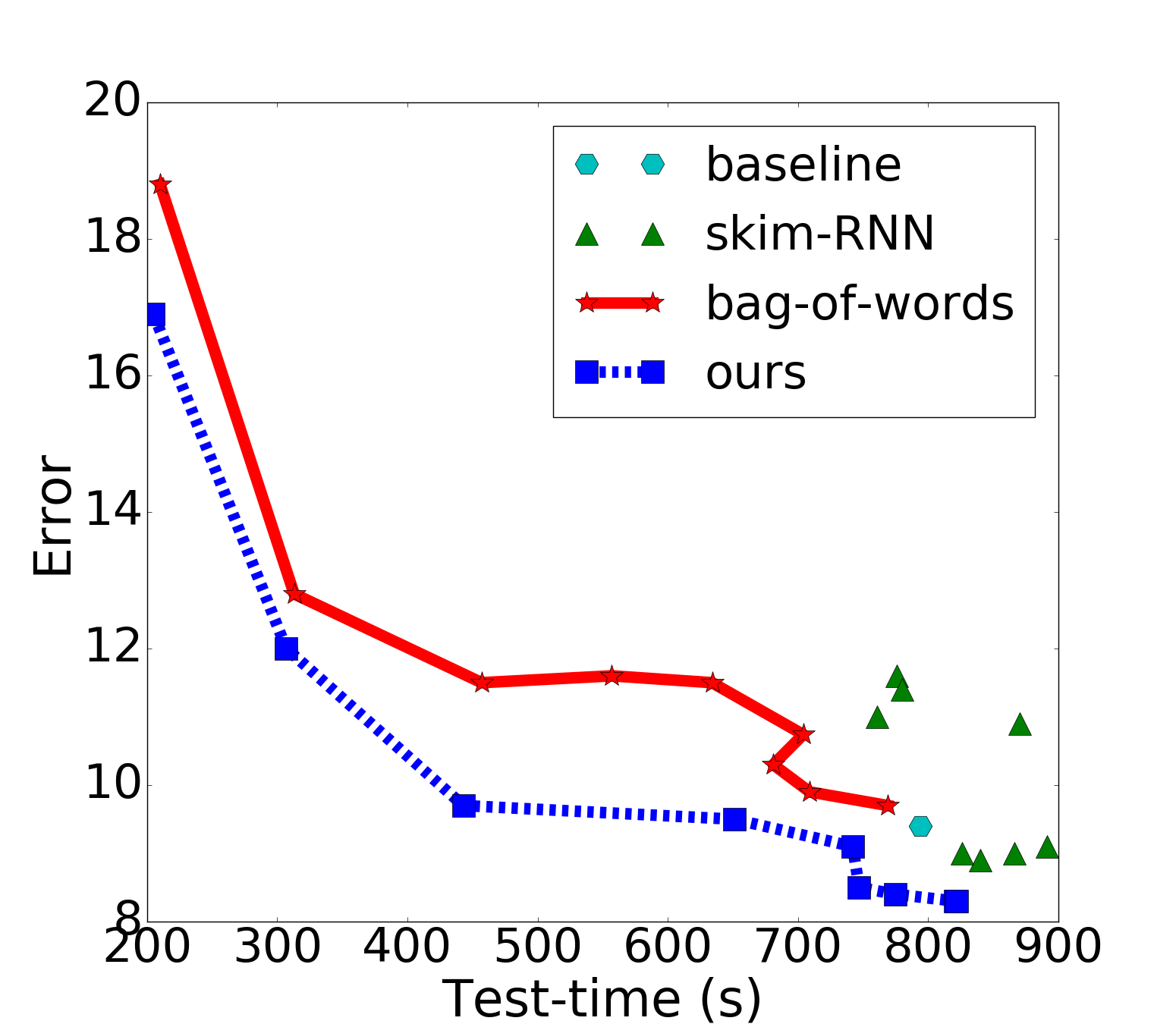}\label{fig:skim-comp-imdb}
        } &
        \hspace{-0.57cm}
        \subfigure[AGNews]{\includegraphics[width=0.374\linewidth,height=0.2770\linewidth]{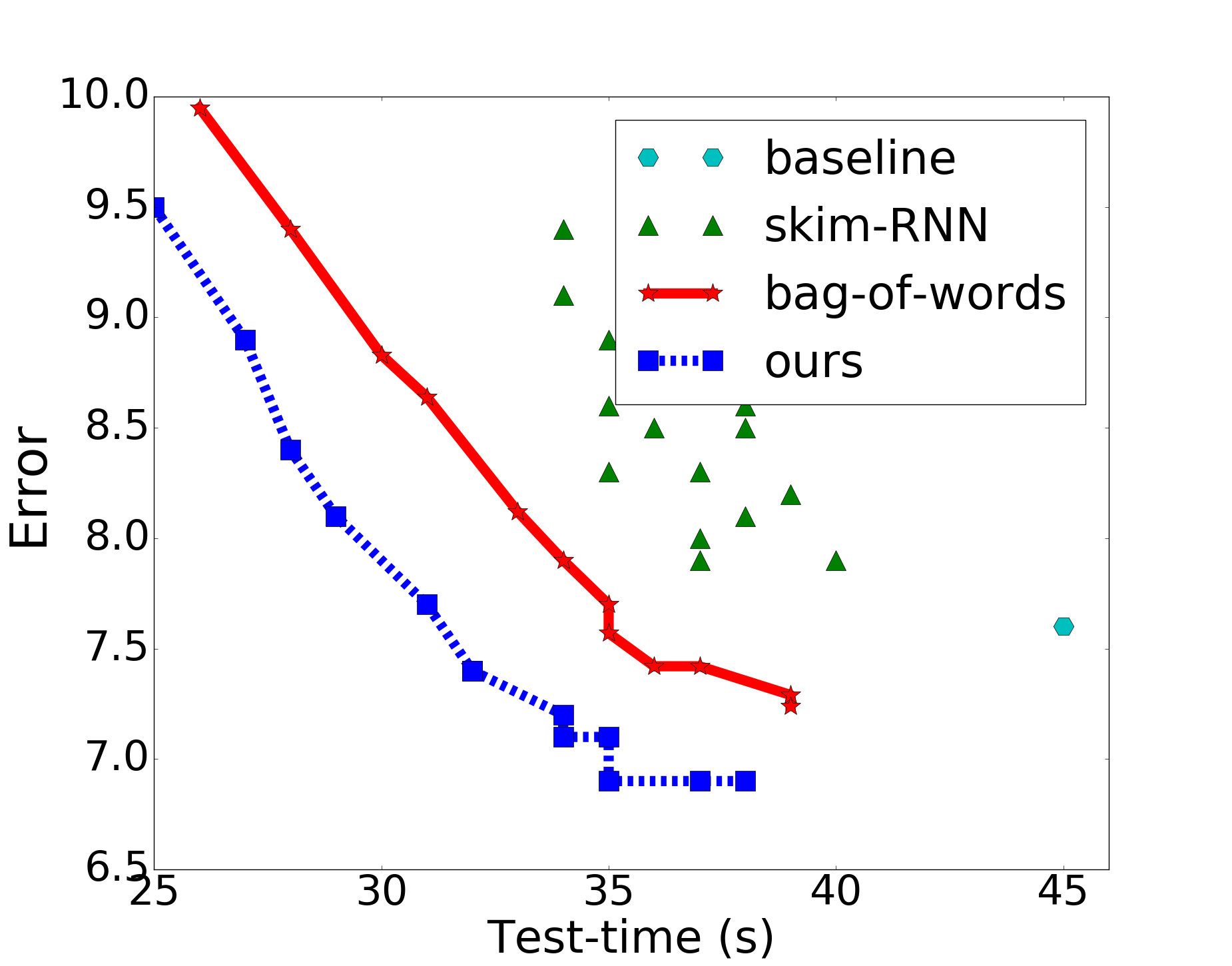} \label{fig:skim-comp-agnews}} 
\hspace{-0.87cm}
  \subfigure[SST-2]{\includegraphics[width=0.374\textwidth, height=0.2770\textwidth]{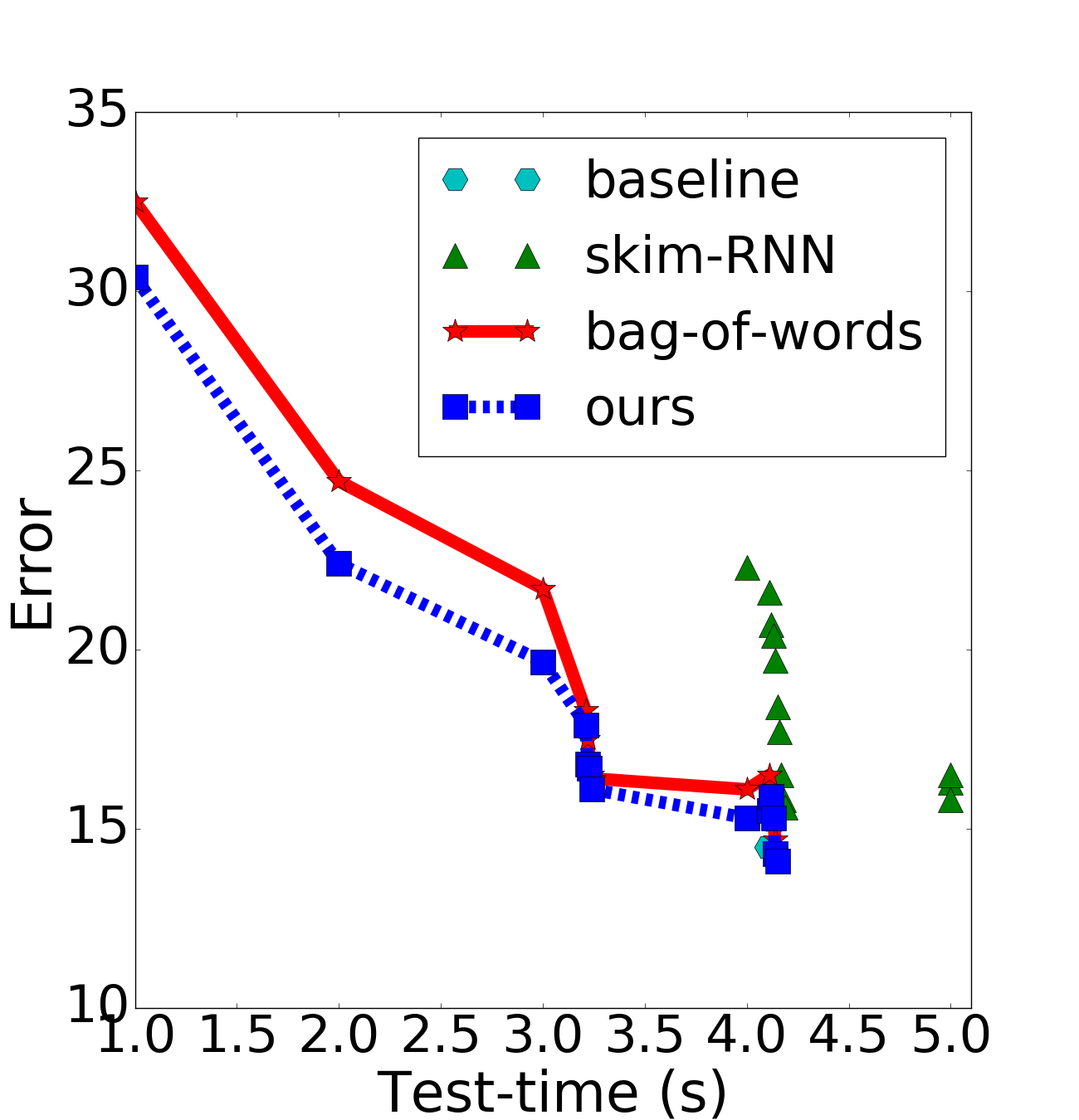}\label{fig:imdb_latm_sel}}
    \end{tabular}
  \caption{ 
  \footnotesize{Performance  under different test-times on IMDB, AGNews, and SST-2. All the approaches use the same LSTM model as the back-end. Bag-of-Words model and our framework have the same bag-of-words {\it selector} cascaded with this LSTM {\it classifier} trained on the original training corpus and aggregated corpus, respectively.
  Our model (blue dashed line) significantly outperform others for any test-time budget. Also its performance is robust, while results of skim-RNN is inconsistent with different budget levels.  
}   
  } 
  \label{fig:tcomp-skim}
\end{figure*}

\begin{table*}[h!]
\centering
\renewcommand{\tabularxcolumn}{m} 
{ 
\footnotesize
\begin{tabularx}{\textwidth}{|c|>{\raggedright}X|}
\hline
 World News & ..  \textbf{plant} searched. Kansai Electric Power's \textbf{nuclear} power \textbf{plant} in Fukui .. \textbf{was searched by police}  Saturday ..
 \tabularnewline
 \hline 
 Business & Telecom Austria taps the Bulgarian {\bf market}. {\bf Telecom} Austria, Austrias {\bf largest} telecoms operator, obtained ..
 \tabularnewline
 \hline
    Sci/Tech & .. Reuters - \textbf{Software security} companies \textbf{and} handset makers, including Finland's \textbf{Nokia} (NOK1V.HE), {\bf are} ..
    \tabularnewline
\hline
\end{tabularx}
}
\caption{ 
\footnotesize{Examples of the WE {\it selector} output on AGNews. Bold words are selected.}
}
\label{tab-qual}
\end{table*}

\section{Experiments}
\label{sec:exp}
To evaluate the proposed approach, 
we consider four benchmark datasets: \emph{SST-2}~\citep{sst}, \emph{IMDB}~\cite{maas}, \emph{AGNews}~\cite{agnews}, and \emph{Yelp} \cite{verydeepconv} and two widely used architectures for classification: \emph{ LSTM},  and \emph{ BCN}~\citep{bcn}. The statistics of the datasets are summarized in Table \ref{data}. 
We evaluate the computation gain of models in terms of overall test-time, and the performance in terms of accuracy. We follow \citet{skim-rnn} to estimate the test-time of models on 
CPU\footnote{Machine specification is in Appendix C.} 
and exclude the time for data loading.

In our approach, we train a {\it classifier} with both WE and bag-of-words {\it selector}s with 6 selection budgets\footnote{For the very large Yelp dataset, 3 selection budgets \{50\%, 60\%, 70\%\} are used.} \{50\%, 60\%, \dots, 100\%\} by the word-level data aggregation framework. We evaluate the computation gain of the proposed method through a comparative study of its performance under different test-times by varying the selection budgets\footnote{In Appendix B, we discuss how to vary these budgets.
} in comparison to the following approaches: (1) \textit{Baseline}: the original {\it classifier} (i.e., no {\it selector}, no data aggregation) (2) \textit{skim-RNN}: we train a skim-RNN model and vary the amount of text to skim (i.e., test-time) by tuning $\theta$ parameter as in \citet{skim-rnn}.  
(3)
\text{Bag-of-Words}: filtering words by the bag-of-words {\it selector} 
and feeding the fragments of sentences to the original {\it classifier} (i.e., no data aggregation). This approach serves as a good baseline and has been considered in the context of linear models (e.g., \citet{chang2008feature}). For a fair comparison, we implement all approaches upon the same framework using AllenNLP library\footnote{\url{https://allennlp.org/}}, including a re-implementation of the existing state-of-art speedup framework skim-RNN~\cite{skim-rnn}\footnote{The official skim-RNN implementation is not released.}. As skim-RNN  is designed specifically for accelerating the LSTM model, we only compare with skim-RNN using LSTM \textit{classifier}. 
Each corresponding model is selected by tuning parameters on validation data. The model is then frozen and evaluated on test-data for different selection budgets.

Figure \ref{fig:tcomp-skim} demonstrates the trade-off between the performance, and the test-time for each setting.
Overall, we expect the error to decrease with a larger test-time budget. 
From Figure \ref{fig:tcomp-skim}, on all of the IMDB, AGNews, and SST-2 datasets, LSTM {\it classifier} trained with our proposed data aggregation not only achieves the lowest error curve but also the results are robust and consistent. That is our approach achieves higher performance across different test-time budgets and its performance is a predictable monotonic function of the test-time budget. However, the performance of skim-RNN exhibits inconsistency for different budgets. As a matter of fact, for multiple budgets, none of the skim-RNN, and LSTM-jump address the problem of different word distribution between training and testing. Therefore, similar to skim-RNN, we anticipate that the behavior of LSTM-jump will be inconsistent as well\footnote{As an example, from Table 6 in \citet{learning-to-skim}, the performance of LSTM-jump drops from 0.881 to 0.854 although it takes longer test-time (102s) than the baseline (81.7s).}. Additionally, since LSTM-jump has already been shown to be outperformed by skim-RNN, we do not further compare with it. 
Next, we show that our framework is generic and can incorporate with other different {\it classifier}s, such as BCN (see Table \ref{tab-bcnresult}).\footnote{Because of the inherent accuracy/inference-time tradeoff, it is difficult to depict model comparisons. For this reason, in Figure 2, we plot  the trade-off curve to demonstrate the best speedup achieved by our model for achieving near state-of-art performance. On the other hand, test results are tabulated in Table 1 to focus attention primarily on accuracy. }
When phrase boundary information is available, our model can further achieve {\bf 86.7} in accuracy with {\bf 1.7x} speedup for BCN on SST-2 dataset by using phrase-level data aggregation. 
Finally, one more advantage of the proposed framework is that the output of the {\it selector} is interpretable.   
In Table \ref{tab-qual}, we present that our framework correctly selects words such as ``Nokia'', ``telecom'', 
and phrases such as ``searched by police'', ``software security'' and filters out words like ``Aug.'', ``users'' and ``products''. 

Note that nevertheless we focus on efficient inference, empirically our method is no more complex than the baseline during training. Despite the number of training instances increases, and so does the training time for each epoch, the number of epochs we require for obtaining a good model is usually smaller. For example, on the Yelp corpus, we only need 3 epochs to train a BCN {\it classifier} on the aggregated corpus generated by using 3 different {\it selector}s, while training on the original corpus requires 10 epochs. 

\section{Conclusion}
\label{conclusion}
We present a framework to learn a robust {\it classifier} under test-time constraints. We demonstrate that the proposed {\it selector}s effectively select important words for {\it classifier} to process and the data aggregation strategy improves the model performance. 
As future work we will apply the framework for other text reading tasks. 
Another promising direction is to explore the benefits of text classification model in an edge-device setting. This problem naturally arises with local devices (e.g., smart watches or mobile phones), which do not have sufficient memory or computational power to execute a complex {\it classifier}, and instances must be sent to the cloud. This setting is particularly suited to ours since we could choose to send only the important words to the cloud. In contrast, skim-RNN and LSTM-jump, which process the text sequentially, have  to either send the entire text to the server or require multiple rounds of communication between the server and local devices resulting in high network latency.

\section{Acknowledgments}
\label{Acknowledgments}
We thank the anonymous reviewers for their insightful
feedback. We also thank UCLA-NLP group for discussion and comments. This work was supported in part by National Science Foundation grants IIS-1760523 and CCF-1527618.

}
\bibliography{emnlp-ijcnlp-2019}
\bibliographystyle{acl_natbib}

\appendix
{
\clearpage
{

\section{Stop-words Removing:}
\label{sec:apndx:stop-words}
Our preliminary experiments show that
although Stop-words achieves notable speedup, it sometimes comes with a significant performance drop. For example, removing Stop-words from SST-2 dataset, the test-time is ~2x faster but the accuracy drops from 85.5 to 82.2. This is due to the stop-words used for filtering text are not learned with the class labels; therefore, some meaningful words (e.g., ``but'', ``not'') are filtered out even if they play a very significant role in determining the polarity of the full sentence (e.g., ``cheap but not healthy''). Besides, we cannot control the budget in the Stop-words approach.

\section{Hyperparameter Tuning:}
\label{tuning}
As the performance is proportionate to the text selected, controlling the selection budget we indeed control the performance. In this section we discuss how to control the selection budget by tuning the hyperparameters.

\subsection{Tuning the WE {\it selector}:} 
For the WE {\it selector}, we vary the selection budget by tuning the two hyperparameters sparsity ($\lambda_1$), and coherent ($\lambda_2)$ of \citet{tao-lei}. In the table below we provide an example settings for corresponding fraction of text to select.
\begin{table}[h]
\centering
\small
  \begin{tabular}{l c r}
    \toprule
      {Sparsity ($\lambda_1)$} & {Continuity ($\lambda_2)$} & {Selection rate (\%)}\\
      \midrule
  8.5e-05 & 2.0 & 2.0 \\
8.5e-05 & 1.0 & 3.0 \\
9.5e-05 & 2.0 & 5.0 \\
9.5e-05 & 1.0 & 6.0 \\
0.0001 & 2.0 & 9.0 \\
0.0001 & 1.0 & 12.0 \\
0.000105 & 2.0 & 13.0 \\
0.000105 & 1.0 & 15.0 \\
0.00011 & 2.0 & 16.0 \\
0.00011 & 1.0 & 22.0 \\
0.000115 & 2.0 & 23.0 \\
0.000115 & 1.0 & 24.0 \\
0.00012 & 2.0 & 28.0 \\
0.00012 & 1.0 & 64.0 \\
    \bottomrule
  \end{tabular}
\end{table}

\newpage

\subsection{Tuning the Bag-of-Words {\it selector}:}
As an example, the following is the regularization hyper-parameter $C$\footnote{\url{https://scikit-learn.org/stable/modules/generated/sklearn.linear_model.LogisticRegression.html}} and corresponding selection rate by the bag-of-words {\it selector} on IMDB.  
\begin{table}[h]
\centering
\small
  \begin{tabular}{l r}
    \toprule
      C & {Selection rate (\%)}\\
      \midrule
0.01 & 27 \\
0.05 & 37 \\
0.1 & 53 \\
0.11 & 63 \\
0.15 & 66\\ 
0.25 & 73\\ 
0.7785 &79\\
1.5 & 82\\
2.5 & 88 \\
      \bottomrule
    \end{tabular}
\end{table}

\subsection{Tuning skim-RNN:}
We re-implement the skim-RNN model as the same baseline as ours  with large RNN size $d=300$, and small RNN sizes $d' \in \{5, 10, 15, 20\}$, and $\gamma \in \{1e^{-9}, 1e^{-10}, 1e^{-11}\}$. For results in Table 2 (in main paper), we compare our model with the best results found from the skim-RNN models with different $d'$, and $\gamma$.  For IMDB, we found the best speedup and accuracy with $d'=10$ and hence for Figure 2 (in main paper), we consider this model with $d'=10$ and vary the selection threshold $\theta$ at inference time as described in ~\citet{skim-rnn} for getting different selection of words. We report the accuracy and the test-time for each setting and plot it in Figure 2 (in main paper).
The following is the selection thresholds for IMDB.
\begin{table}[h]
\centering
\small
  \begin{tabular}{l r}
    \toprule
      $\theta$	& skimmed(\%)\\
      \midrule
0.45 & 	99\\ 
0.48 &	97 \\
0.47 &	93 \\
0.505 &	63  \\
0.51 &	54 \\
0.52 &	34 \\
0.53 &	20 \\ 
      \bottomrule
    \end{tabular}
\end{table}

\clearpage


\section{Machine Specification:}
\label{spec}
\begin{lstlisting}
Architecture:          x86_64
CPU op-mode(s):        32-bit, 64-bit
Byte Order:            Little Endian
CPU(s):                12
On-line CPU(s) list:   0-11
Thread(s) per core:    2
Core(s) per socket:    6
Socket(s):             1
NUMA node(s):          1
Vendor ID:             GenuineIntel
CPU family:            6
Model:                 63
Stepping:              2
CPU MHz:               1200.890
BogoMIPS:              6596.22
Virtualization:        VT-x
L1d cache:             32K
L1i cache:             32K
L2 cache:              256K
L3 cache:              15360K
NUMA node0 CPU(s):     0-11
\end{lstlisting}

}



%

}

\end{document}